\PassOptionsToPackage{utf8}{inputenc}
\documentclass{bioinfo}
\copyrightyear{2022} \pubyear{2022}

\usepackage[hidelinks]{hyperref}
\usepackage{float}
\usepackage{graphicx}
\usepackage{subfigure}
\usepackage{booktabs}

\access{}
\appnotes{}

\begin{document}
\firstpage{1}

\subtitle{System biology}

\title[NIAPU]{NIAPU: network-informed adaptive positive-unlabeled learning for disease gene identification}
\author[Stolfi \textit{et~al}.]{Paola Stolfi\,$^{\text{\sfb 1}}$, Andrea Mastropietro\,$^{\text{\sfb 2,}*}$, Giuseppe Pasculli\,$^{\text{\sfb 2}}$, Paolo Tieri\,$^{\text{\sfb 1}}$ and Davide Vergni\,$^{\text{\sfb 1,}*}$}
\address{$^{\text{\sf 1}}$Institute for Applied Computing (IAC) “Mauro Picone”, National Research Council of Italy (CNR), Rome, 00185, Italy and \\
$^{\text{\sf 2}}$Department of Computer, Control and Management Engineering (DIAG) “Antonio Ruberti”, Sapienza University of Rome, 00185, Italy.}

\corresp{$^\ast$To whom correspondence should be addressed.}

\history{}

\editor{}

\abstract{\textbf{Motivation:} Gene-disease associations are fundamental for understanding disease etiology and developing effective interventions and treatments. Identifying genes not yet associated with a disease due to a lack of studies is a challenging task in which prioritization based on prior knowledge is an important element. The computational search for new candidate disease genes may be eased by positive-unlabeled learning, the machine learning setting in which only a subset of instances are labeled as positive while the rest of the data set is unlabeled. In this work, we propose a set of effective network-based features to be used in a novel Markov diffusion-based multi-class labeling strategy for putative disease gene discovery.\\
\textbf{Results:} The performances of the new labeling algorithm and the effectiveness of the proposed features have been tested on ten different disease data sets using three machine learning algorithms. The new features have been compared against classical topological and functional/ontological features and a set of network- and biological-derived features already used in gene discovery tasks. The predictive power of the integrated methodology in searching for new disease genes has been found to be competitive against state-of-the-art algorithms.\\
\textbf{Availability and implementation:} The source code of NIAPU can be accessed at \url{https://github.com/AndMastro/NIAPU}. The source data used in this study are available online on the respective websites.\\
\textbf{Contact:} \href{mailto:mastropietro@diag.uniroma1.it}{mastropietro@diag.uniroma1.it}, \href{mailto:davide.vergni@cnr.it}{davide.vergni@cnr.it}.}

\maketitle

\section{Introduction}
\label{sec:intro}
The discovery of gene-disease associations (GDAs) is made difficult by incomplete knowledge of biological and physiological processes. When approaching complex, multi-gene diseases and traits, it is hard to disentangle the contribution of each gene, and computational biological approaches for predicting GDAs \citep{Opap2017, Piro2012} can support and address experimental methods (e.g., Genome-Wide Association Studies -- GWAS -- or linkage studies, among others) which are expensive and time-consuming.

The fuzzy background of yet unknown or truly unassociated genes contributes to making the computational identification of disease genes challenging to carry out with accuracy.
In machine learning (ML), this setting translates into the ability to identify new positive instances among a set of positive and unlabeled samples, a task known as “positive-unlabeled (PU) learning” \citep{liu2003,Bekker2020}.
This task can be addressed through semi-supervised learning algorithms, trained using two approaches. In the first one, the set of unlabeled instances is assumed to be a contaminated set of negative instances and the contamination is considered during the modeling process by weighting the data points or adding penalties on misclassification \citep{elkan2008, mordelet2014, claesen2015, ke2018}. In the specific case of gene discovery, this contamination is given by the possibility of the negative instances of containing not yet discovered positive genes. The second approach, called two-step technique, aims at relabeling the instances and then training a supervised learning algorithm \citep{liu2003, Yang2012, Yang2014}. For example, \cite{Yang2012} introduced a multi-class labeling procedure considering five different labels, namely Positive (P), Likely Positive (LP), Weakly Negative (WN), Likely Negative (LN), and Reliable Negative (RN), based on a Markov process with restart \citep{can2005}, widely applied in disease genes identification \citep{kohler2008, li2010, li2010genome}. Then, a supervised learning algorithm is trained on the relabeled data.

In the present work, we considered the multi-class labeling approach since it allows identifying a set of originally unlabeled items, namely the LP set, whose features are close to that of the items in P. This translates into the identification of a small set of genes more likely to contain true positive instances, hence providing a set of new candidate disease genes for prioritization.

Going beyond the approach from \cite{Yang2012}, we propose several significant modifications of the multi-class method regarding the distance matrix defining the Markov process and the selection of the different classes. Some of these modifications were needed in order to apply the method to general PU data sets, while others were proposed to make the process of class formation more rigorous and, at the same time, flexible.
The approach considered here, being a two-step technique, is based on the separability and smoothness assumptions \citep{Bekker2020}, which require that the features should be able to distinguish between positive and negative instances and, at the same time, instances with similar features should be more likely to have the same label. Therefore, as a further contribution, we propose the use of specific network-informed features, one of them introduced for the first time in this work, based on protein-protein interaction (PPI) data, which provide a characterization of the topological relationships of all human genes with respect to the set of
disease genes. The use of such measures grants a much more precise classification of genes than other topological measures. In particular, the set of seed genes is identified very precisely as well as the genes closest and farthest to them, as shown in Section~\ref{subsec:5foldCV}. The Network-Informed Adaptive Positive-Unlabeled (NIAPU) framework is therefore formed by two components: the Network Diffusion and Biology-Informed Topological (NeDBIT) features and the Adaptive Positive-Unlabeled (APU) labeling algorithm.
\begin{methods}
\section{Materials and methods}
\label{sec:method}
\subsection{Data sources and preprocessing}
\label{subsec:data}
The proposed methodology exploits two types of data, i.e., reliable PPIs and known GDA data.
PPI data provide valuable biological knowledge for the identification of undiscovered disease genes \citep{Piro2012,Doncheva2012,Silverman,TIERI2019805, petti2021moses}. Human PPI data, i.e., the human interactome, were gathered from the BioGRID \citep{stark2006biogrid} dataset\footnote{Version 4.4.206.}. The human interactome is obtained by choosing Homo sapiens genes (organism ID 9606), from which we extract a connected network consisting of 19761 genes and 678932 nonredundant, undirected interactions (see Supplementary File 1).

GDAs were derived from DisGeNET\footnote{Version 7.0.} \citep{pinero2016disgenet, pinero2020disgenet}, a discovery platform containing one of the largest publicly available collections of genes and variants associated with human diseases together with a score denoting the association confidence and significance. Ten diseases were considered: malignant neoplasm of breast (disease ID C0006142, 1074 genes), schizophrenia (C0036341, 883 genes), liver cirrhosis (C0023893, 774 genes), colorectal carcinoma (C0009402, 702 genes), malignant neoplasm of prostate (C0376358, 616 genes), bipolar disorder (C0005586, 477 genes), intellectual disability (C3714756, 447 genes), drug-induced liver disease (C0860207, 404 genes), depressive disorder (C0011581, 289 genes), and chronic alcoholic intoxication (C0001973, 268 genes). The selection criterion for these diseases was the highest cardinality of GDAs in the DisGeNET curated dataset to ensure sufficient information for the ML task. To validate the gene discovery results, we relied on the \emph{all genes} DisGeNET dataset, which we refer to as \emph{extended} dataset. The latter contains associated genes from additional sources not present in the curated version \citep{bundschus2008extraction,bundschus2010digging, bravo2014knowledge,bravo2015extraction}. More details can be found in Supplementary File 2. After performing additional cleaning steps (see Supplementary File 2), we ended up having a set of seed genes for each disease, denoted by $\Sigma$, with their association score $\mathbf{S}$. In particular, we have 1025 genes for disease C0006142, 832 for C0036341, 747 for C0023893, 672 for C0009402, 606 for C0376358, 451 for C0005586, 431 for C3714756, 320 for C0860207, 279 for C0011581, and 255 for C0001973.

\subsection{Multi-class labeling: Adaptive PU (APU) labeling algorithm and classification}\label{sec:APU}

The APU algorithm consists of a multi-class labeling procedure that relies on the labels introduced in \citealp{Yang2012}: P, LP, WN, LN, and RN. P instances are the known disease genes, RN instances represent the genes whose features are the furthest from the average features in the P set, while the remaining labels are assigned through a Markov process with restart \citep{can2005}.
The novelty of the proposed method is the construction of a new transition matrix starting from the distance matrix between the features of the genes. The matrix needs to be normalized in order to preserve the total transition probability of the state vector whose initial value is different from zero only for genes in the P and RN classes. Moreover, the class selection has been made flexible by using an adaptable quantile separation instead of fixed thresholds. These characteristics have been implemented in order to make the process of class formation more rigorous and, at the same time, more flexible hence easily adaptable to different settings, datasets, and needs.

Let $V$ be a set whose generic $i^{th}$ element $v_{i=1,...,n}$ is characterized by the couple $\left(\boldsymbol{x}_{i},y_{i}\right)$ where $\boldsymbol{x}_{i}\in \left[0,1\right]^{d}$ represent the feature vector, and $y_{i}\in \left\{0,1\right\}$ the initial label. 
The APU algorithm is defined by the following steps:\\
{\it \textbf{Step 1}:}
Compute the matrix $\boldsymbol{W}$, whose elements $w_{ij}$ are defined as follows
\begin{equation}
w_{ij} = \begin{cases}1- \frac{e_{ij}-m}{M-m} &\mbox{if } i\neq j\\ 1 & \mbox{otherwise} 
\end{cases} 
\label{eq:matrix}
\end{equation}
where $e_{ij}=\sum_k \left (x^k_i-x^k_j\right)^2$, $m=\mbox{min}_{ij}\left\{e_{ij}\right\}$ and $M=\mbox{max}_{ij}\left\{e_{ij}\right\}$. The symmetric matrix $\boldsymbol{W}$ represents the similarity score between elements $i$ and $j$.\\
{\it \textbf{Step 2}:}
Compute the reduced matrix $\boldsymbol{W}_{r}$ as follows 
\begin{equation}
w_{r, ij}=\begin{cases}w_{ij} &\mbox{if } w_{ij}>q_{\boldsymbol{w}}\\ 0 & \mbox{otherwise} 
\end{cases}. \nonumber
\end{equation}
The threshold $q_{\boldsymbol{w}}$ is computed as a given quantile of the distribution of the elements in the matrix $\boldsymbol{W}$ in order to exclude from the propagation process links between poorly related elements. To obtain a proper Markov process, i.e., preserving the probability distribution, the matrix $\boldsymbol{W}_{r}$ must be normalized as
$\boldsymbol{W}_{n}= \boldsymbol{D}^{-1}\boldsymbol{W}_{r}$,
where $\boldsymbol{D}$ is the diagonal matrix with elements $d_{ii}=\sum_{j}w_{r, ij}$.\\
{\it \textbf{Step 3}:}
Initialize the propagation process with the initial state vector $\boldsymbol{g}_{0}$ defined as follows. Let $\vert P\vert$ be the cardinality of P (set of seed genes) and $\boldsymbol{\hat x}=\left(\hat{x}^{1}, \dots, \hat{x}^{d}\right)$, where ${\hat x}^k=1/|P|\sum_{i\in P}x^k_i$, be the average features of P. The RN genes are chosen to be the ones having the most distant features from $\boldsymbol{\hat x}$. We select the $|P|$ most distant genes from $\boldsymbol{\hat x}$ in order to keep the classes balanced. 
Then, the $i$-th element of $\boldsymbol{g}_{0}$ is defined as 
\begin{equation}
g_{0,i} =\begin{cases}
1 \qquad\mbox{if}\qquad i\in P\\ 
-1 \quad\mbox{if}\qquad i\in RN\\
0 \qquad\mbox{otherwise}
\end{cases}. \nonumber
\end{equation}
When needed, a different number of RN genes can be selected. In this case, the initial value of the RN genes in the state vector $\boldsymbol{g}_{0}$ must be set to $-\vert P\vert/\vert RN\vert$ so that the two distributions of positive and negative values are balanced in $\boldsymbol{g}_{0}$, with the sum of its elements equal to zero.\\
{\it \textbf{Step 4}:} Define a Markov process with restart as 
\begin{equation}
\label{eq:mpr}
\boldsymbol{g}_{r}=\left(1-\alpha\right)\boldsymbol{W_n^t}\boldsymbol{g}_{r-1}+\alpha \boldsymbol{g}_{0},
\end{equation}
where the parameter $\alpha$ is usually set to $0.8$ \citep{Yang2012,li2010genome}. 
Starting from the state vector $\boldsymbol{g}_0$ the dynamics in Equation~\eqref{eq:mpr} ends in the stationary state $\boldsymbol{g}_{\infty}$, numerically reached when $\vert \boldsymbol{g}_{r}-\boldsymbol{g}_{r-1}\vert < 10^{-6}$. \\
{\it \textbf{Step 5}:} Use $G_{\infty}$ to assign the remaining labels. Selecting only the elements that belong neither to P nor to RN, the values of the asymptotic distribution of those elements are sorted and the ranking of the corresponding elements is used to form the remaining classes: LP, WN and LN. A simple rule is to divide the ranking into three equal parts and identify LP samples with the first third, WN with the second third and LN with the third third. However, depending on the type of analysis and the problem addressed, any division of the ranking can be considered acceptable.\\
{\it \textbf{Step 6}:} Classification. An ML classifier is trained over the data set containing features and labels. Three different ML algorithms have been used: Random Forest (RF) \citep{breiman2001}, Support Vector Machine (SVM) \citep{cortes1995,drucker1997} and Multilayer Perceptron (MLP) \citep{hastie2001} (details in Supplementary File 2).

\subsection{Network Diffusion and Biology-Informed Topological (NeDBIT) features}\label{sec:features}
The NeDBIT features include two network diffusion-based features, namely heat diffusion and balanced diffusion, and two biology-informed topological metrics, namely NetShort and NetRing. Network diffusion methods are widely used in disease gene discovery processes \citep{Lancour,Picart,Janyasupab}. We coupled network diffusion methods and innovative topological-based features in order to make the most of the combined predictive power of both approaches. 
Moreover, all the features are computed exploiting the association score $\mathbf{S}$. In this way, the NeDBIT features, not assigning the same weight to all seed genes, are certainly more significant for the disease under investigation.\\

\noindent
{\bf Heat diffusion feature}\\
This feature is obtained by using a heat diffusion process over the network, which is among the most used processes for disease gene prioritization and prediction (see \citealp{diffusioncyto} and references therein). Starting with a distribution of weights, with positive values only on the seed genes, their evolution is determined by using the diffusion equation on graph \citep{pmid20840752}
\begin{equation}
{\mathbf z}'(t)+{\mathbf L} {\mathbf z}(t)=0\,,
\label{eqheatdiff}
\end{equation}
where ${\mathbf L}$ is the Graph Laplacian matrix, ${\mathbf L}={\mathbf K} - {\mathbf A}$, $\boldsymbol{K}$ is the diagonal matrix with the degree of nodes on the diagonal, namely ${\mathbf K}_{ii} = k_i$, and ${\mathbf A}$ is the adjacency matrix of the PPI. The weights at time $t$ are given by the formal solution of Equation~(\ref{eqheatdiff})
\begin{equation}
{\mathbf z}(t)=\exp\left ( -{\mathbf L} t \right){\mathbf z}(0),
\label{solheatdiff}
\end{equation}
where $\exp$ is the exponential of the matrix. Regarding the initial distribution of weights, we assign $z_i(0) = s_i$ for seed genes in $\Sigma$ and 0 otherwise, where $s_i$ is the association score.\\ 

\noindent
{\bf Balanced diffusion feature}\\
This feature is obtained by using the diffusion equation in \eqref{eqheatdiff} but with another version for the Graph Laplacian matrix, i.e., ${\mathbf L_b}={\mathbf I} - {\mathbf K}^{-1}{\mathbf A}$. The weights at time $t$ are obtained as in Equation~\eqref{solheatdiff} by using operator ${\mathbf L_b}$ and the initial weights are given as for the previous measure.

This form of the graph diffusion operator differs from the heat diffusion in the fact that the operator ${\mathbf L}$ diffuses the same amount of score for each link, whereas ${\mathbf L_b}$ diffuses the same amount of score for each node. This implies a different short-time behavior of the diffusion process on the graph. \\

\noindent
{\bf NetShort}\\
The NetShort measure \citep{white2003algorithms} is based on the idea that a generic node is topologically important for a disease if a large number of seed nodes must be traversed to reach it.
For each node, the weights are assigned as follows 
\begin{equation}
w_{ij}=a_{ij} \frac{2}{\tilde s_i+\tilde s_j}, \quad \mbox{where} \quad \tilde s_i =\begin{cases}
\frac{s_{i}}{\max{\mathbf{S}}} \qquad\mbox{if}\qquad i\in \Sigma\\ 
		\alpha  \frac{\min{\mathbf{S}}}{\max{\mathbf{S}}} \quad\mbox{if}\qquad i\notin \Sigma
		\end{cases}\nonumber
\end{equation}
and $\min{\mathbf{S}}$ and $\max{\mathbf{S}}$ are the minimum and the maximum of the association scores, $\alpha$ is the penalization parameter given to non-seed nodes, and $a_{ij}$ is the $(i,j)$ element of the adjacency matrix $\boldsymbol{A}$. We use $\alpha=0.5$ so that all non-seed nodes have normalized score
$\tilde s_i =\frac{1}{2}\frac{\min \mathbf{S}}{\max \mathbf{S}}$ while seed nodes have normalized score $\frac{\min \mathbf{S}}{\max \mathbf{S}}\leq \tilde s_i \leq 1$.
Then, the NetShort measure $NS_{i}$ of node $i$ is defined as
\begin{equation}
NS_{i}=\sum_{j\neq i} \frac{1}{d_{ij}},
\nonumber
\end{equation}
where $d_{ij}$ is the length of the weighted shortest path from $i$ to $j$.\\

\noindent
{\bf NetRing}\\
The NetRing measure, introduced for the first time in this work, is based on the concept of ring structure \citep{baronchelli2006ring} generalized to a set of seed nodes. Starting from seed nodes, a partition of the graph in sub-graphs, or rings, is introduced with the following property
\begin{equation}
R(l)\equiv \left \{j\in V \,\,\,|\,\,\,\min_{i\in \Sigma}l_{ij}=l\right \},\nonumber
\end{equation}
where $l_{ij}$ is the (unweighted) length of the shortest path from $i$ to $j$. $R(l)$ contains all the non-seed nodes with a minimal distance $l$ from, at least, one seed node. From the definition follows that $R(0)\equiv\Sigma$, $R(l_1)\cap R(l_2)={\emptyset}$ if $l_1\neq l_2$ and $V=\cup_{l=0}^{L} R(l)$, where $L$ is the highest value of the minimal distance from non-seed nodes to seed nodes.

An initial rank defined by means of the association score is computed as
\begin{equation}
    \hat r_i=\begin{cases}
    1 - \frac{s_i}{\max{\mathbf{S}}} \qquad\mbox{if}\qquad i\in \Sigma\\
    1 \phantom{- \frac{s_i}{\max{s}} \qquad} \ \ \mbox{if}\qquad i\notin \Sigma
    \end{cases}\nonumber,
\end{equation}
then the NetRing measure $r_{i}$ of node $i$ is defined as
\begin{equation}
    r_i=\begin{cases}
    \displaystyle 
    \alpha \hat r_i + 
         (1-\alpha) \frac{1}{k_i}\sum_{j|A_{ij}\neq 0}\hat r_j \qquad \qquad \quad \hspace{1.3cm} \mbox{if}\quad i\in \Sigma \vspace{0.3cm}\\
    \displaystyle 
    l_{i} + \frac{1}{k_i}\left(\sum_{j\in O_i} \hat r_j+ \sum_{j\in R_{i}(l_{i}-1)}r_j -(l_{i}-1)
 \right) \ \mbox{if}\quad i\notin \Sigma
    \end{cases}\nonumber,
\end{equation}
where the score for seed genes is a convex combination of the initial rank $\hat r_i$ and the average of the initial rank of the neighbors of the node, so that seed nodes having many seed nodes as neighbors have a higher rank. The rank of non-seed nodes is obtained by summing the level of the ring and the average of two terms, i.e., the number of genes belonging to the same or higher rings ($O_i=\left \{j\notin R(l-1) | A_{ij}\neq 0\right\}$) and the sum of the rank of genes in the lower ring ($R_i(l_{i}-1)=\left \{j\in R(l_{i}-1) | A_{ij}\neq 0\right\}$) corrected by the ring level. The correction is introduced to make the rank $r_j$ comparable with $\hat r_j$.
Additional important considerations about the NetRing measure can be found in Supplementary File 2.

\end{methods}
\section{Results}
\label{sec:results}
\begin{figure*}
    \centering
    \includegraphics[width=\textwidth]{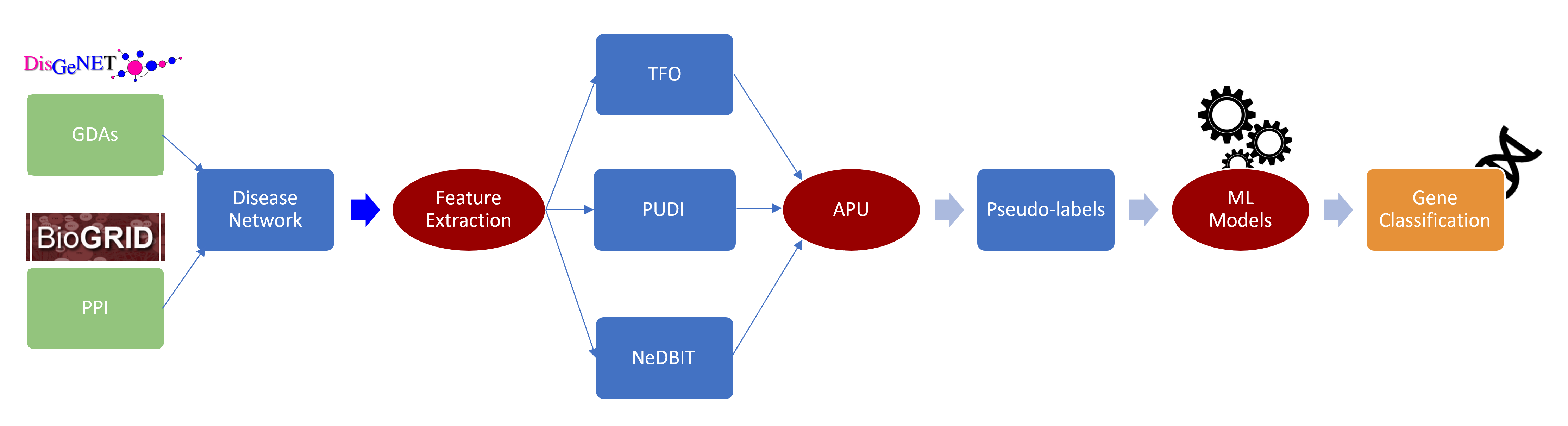}
    \caption{The complete NIAPU pipeline. PPI and GDAs are used to obtain a disease-related network. Features are extracted (Section~\ref{sec:features}) and APU is applied (Section~\ref{sec:APU}) to assign new labels to train ML algorithms for the final gene classification. The new labels can be used for disease gene-discovery purposes (Section~\ref{sec:gene-discovery}).}
    \label{fig:NIAPU-workflow}
\end{figure*}

The performance of NIAPU is tested on the ten disease datasets detailed in Section~\ref{subsec:data}. A visual overview of the workflow can be grasped in Figure~\ref{fig:NIAPU-workflow}. 
Section \ref{subsec:5foldCV} is devoted to testing the performance of NIAPU (APU+NeDBIT) against the implementation of the APU labeling algorithm with two different sets of features commonly used when dealing with disease gene identification. The performances are investigated in terms of out-of-sample classification. Section~\ref{subsec:LP} analyzes the performance of NIAPU in the identification of candidate disease genes. To this end, a subset of seed genes is masked out to see whether such genes are predicted as LP. Section~\ref{sec:gene-discovery} deals with comparing NIAPU with other disease gene identification algorithms, while Section~\ref{sec:enrichment} presents results from an enrichment analysis of the candidate disease genes obtained by the NIAPU methodology.

\subsection{NeDBIT classification performances}
\label{subsec:5foldCV}
The effectiveness of the NeDBIT features is tested by comparing NIAPU against the implementation of the APU labeling algorithm with two different sets of features: the first (PUDI) computed following \citealp{Yang2012} is based on topological features (originally taken from \citealp{Xu2006}) and functional information based on the semantic similarity of GO terms (originally taken from \citealp{wang2007new}), the second (TFO) includes simple topological, functional, and ontological features (see Supplementary Files 2 and 3).
The comparison is carried out in terms of out-of-sample classification performance, namely the ten datasets detailed in Section~\ref{subsec:data} were split into training set (70\%) and test set (30\%), keeping class balance. Then, we trained the three ML algorithms defined in Step 6 of Section~\ref{sec:APU} for the three different applications of the APU algorithm.

\begin{figure*}
    \centering
    \subfigure[MLP + TFO features]{\includegraphics[width=0.32\textwidth]{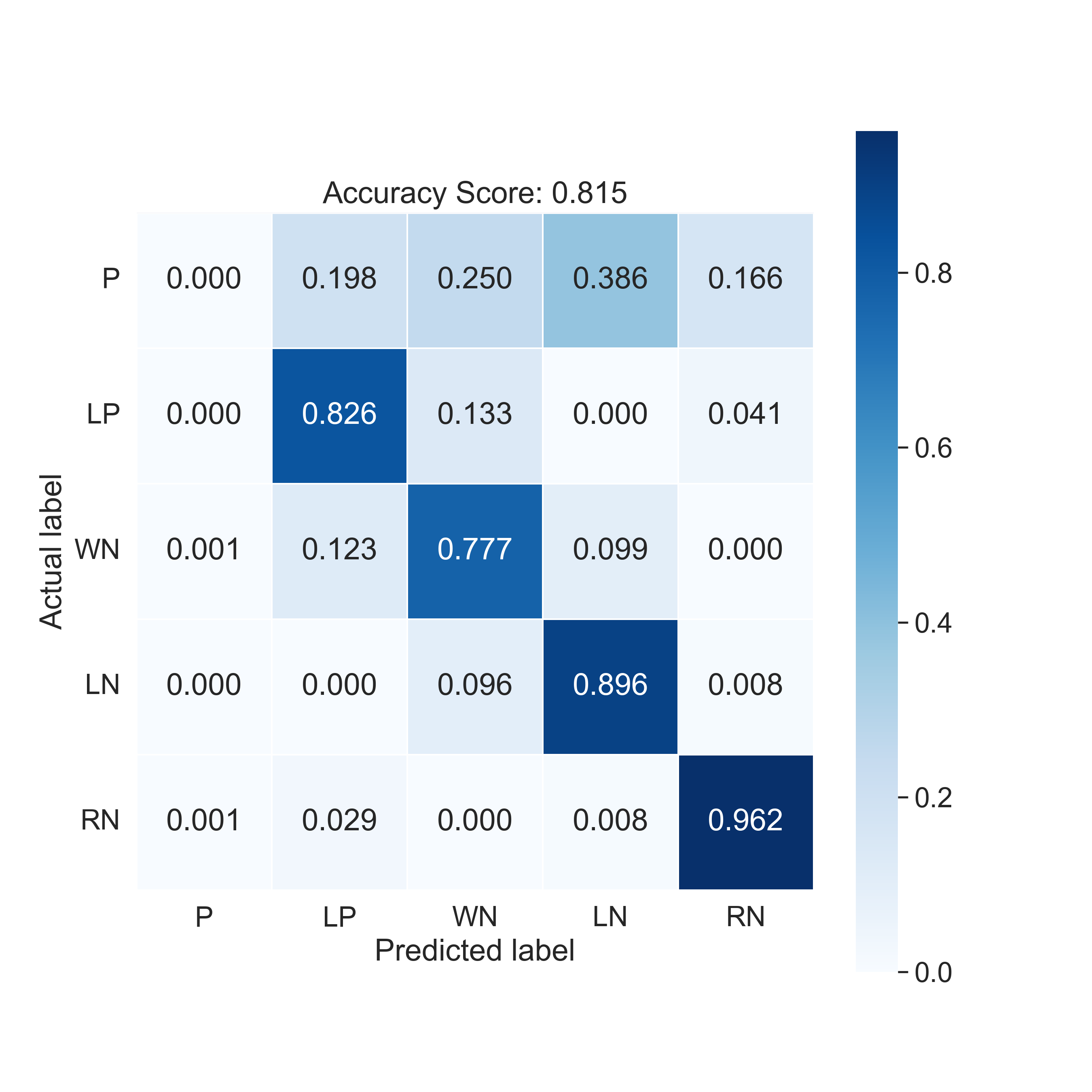}} 
    \subfigure[MLP + PUDI features]{\includegraphics[width=0.32\textwidth]{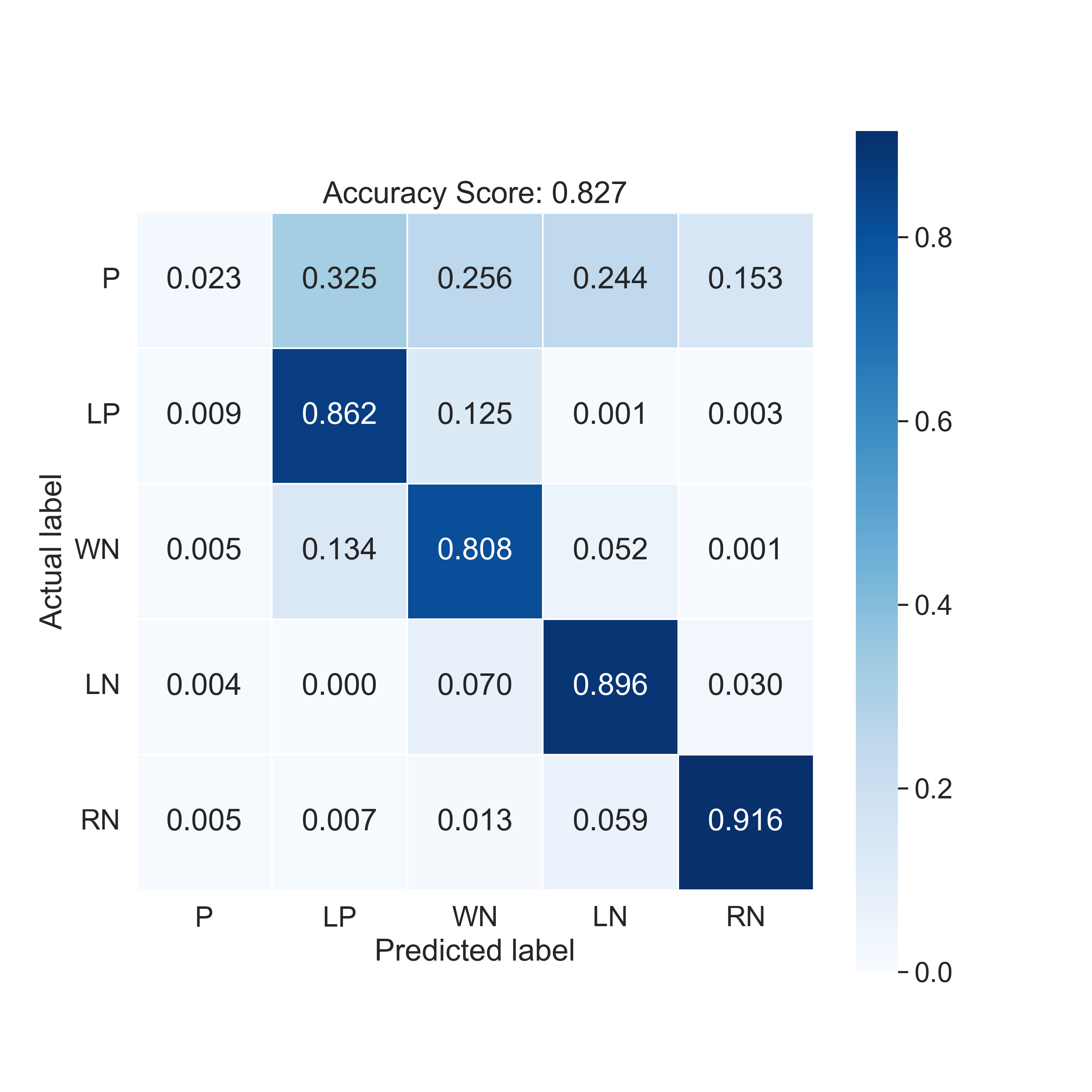}} 
    \subfigure[MLP + NeDBIT features]{\includegraphics[width=0.32\textwidth]{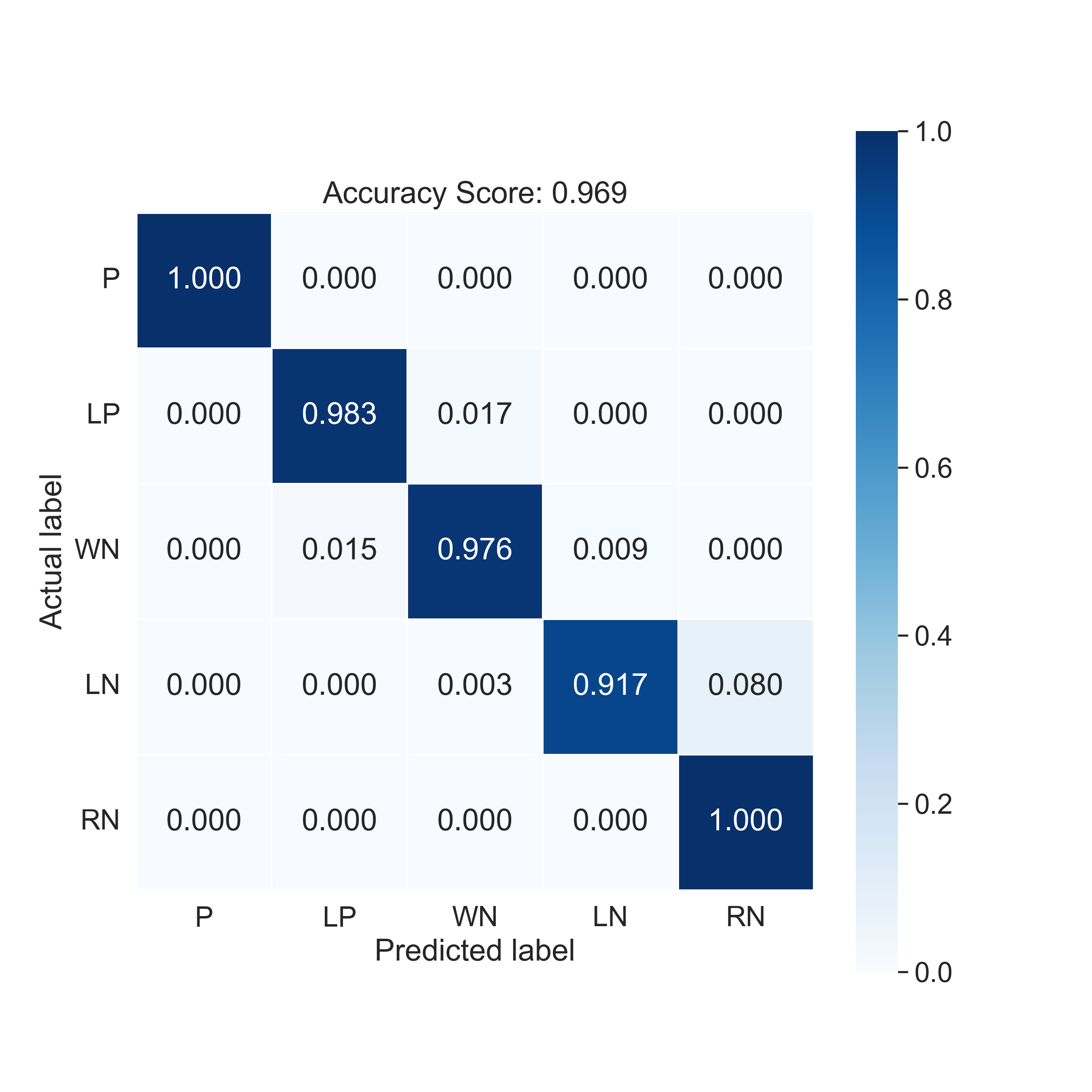}}\\
    
    \subfigure[RF + TFO features]{\includegraphics[width=0.32\textwidth]{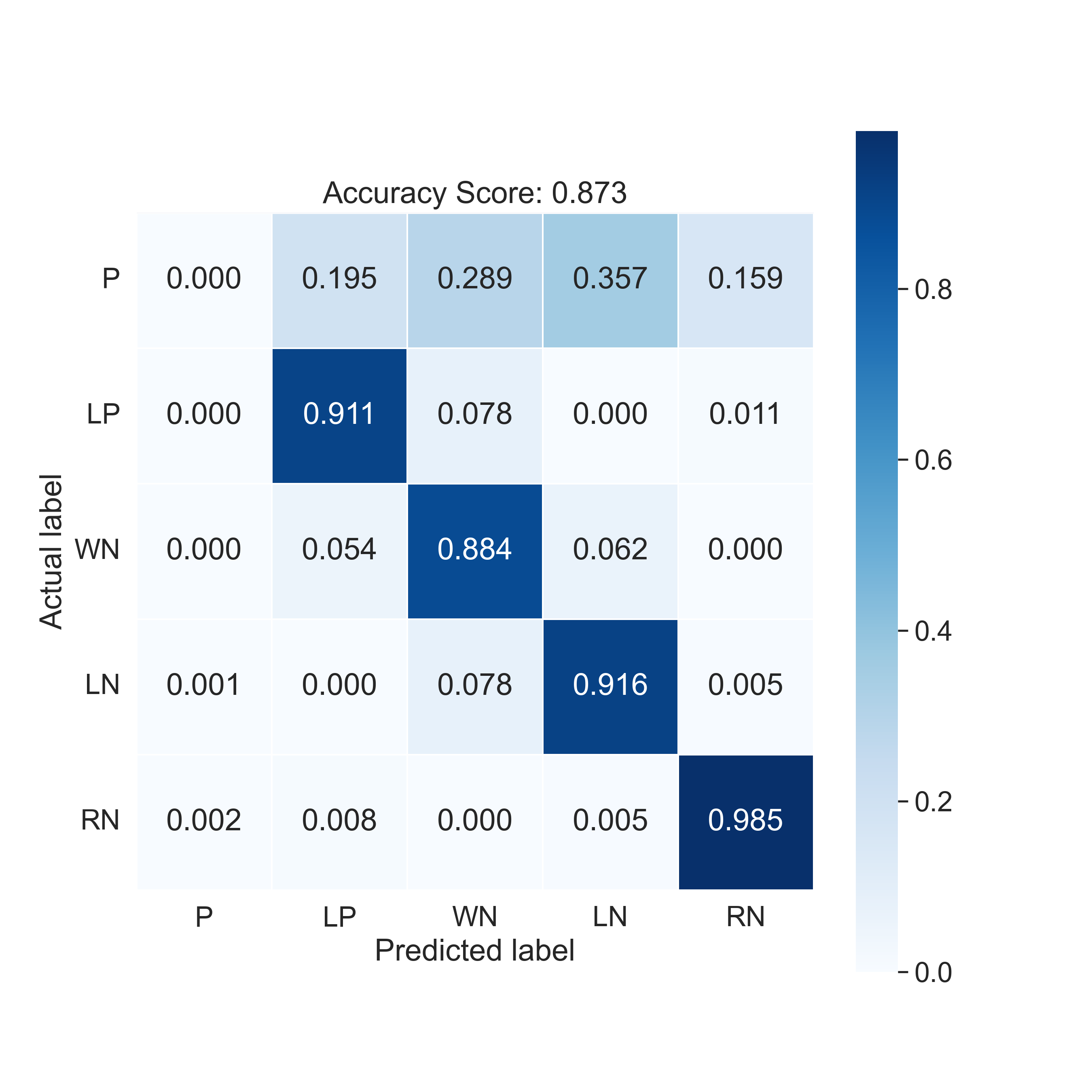}} 
    \subfigure[RF + PUDI features]{\includegraphics[width=0.32\textwidth]{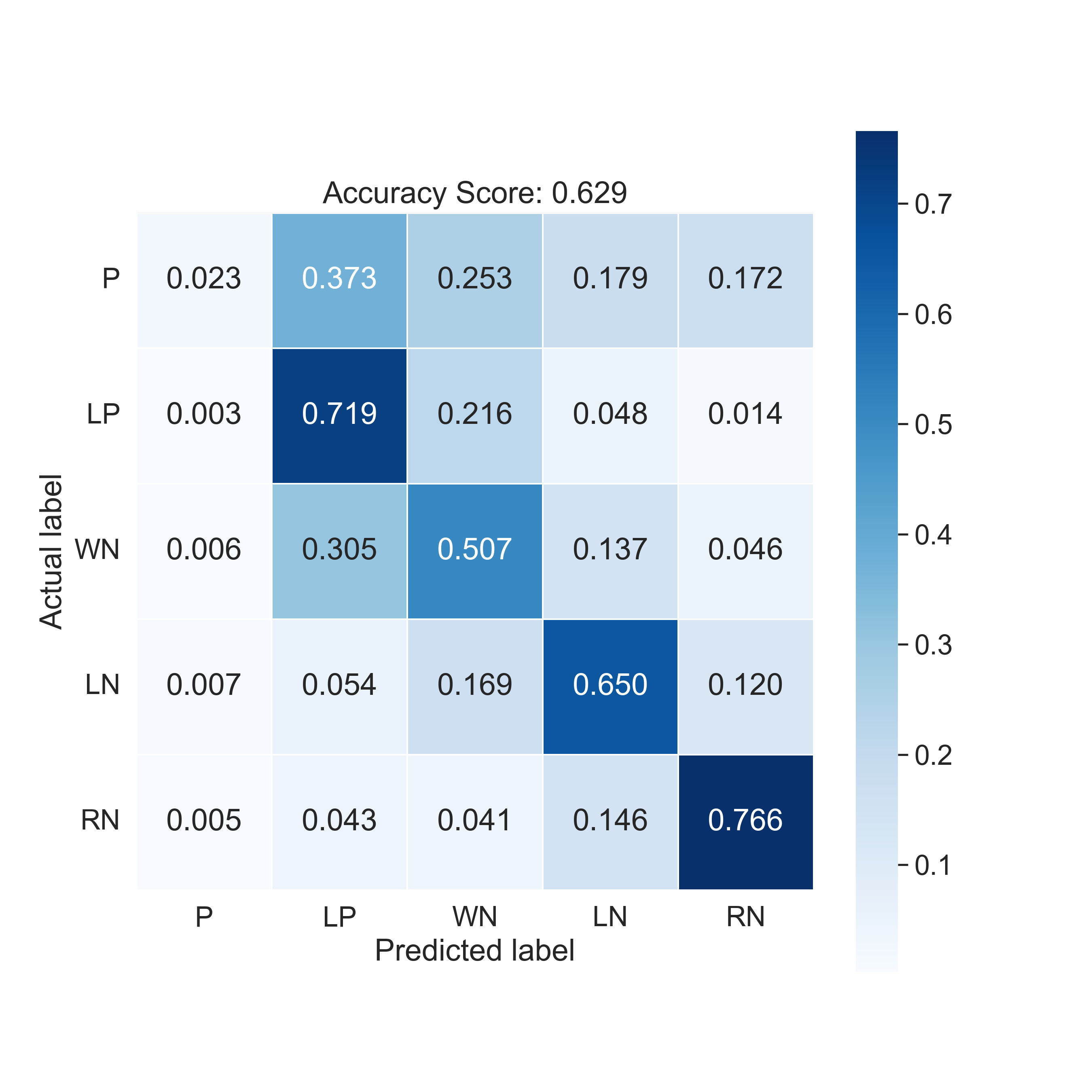}} 
    \subfigure[RF + NeDBIT features]{\includegraphics[width=0.32\textwidth]{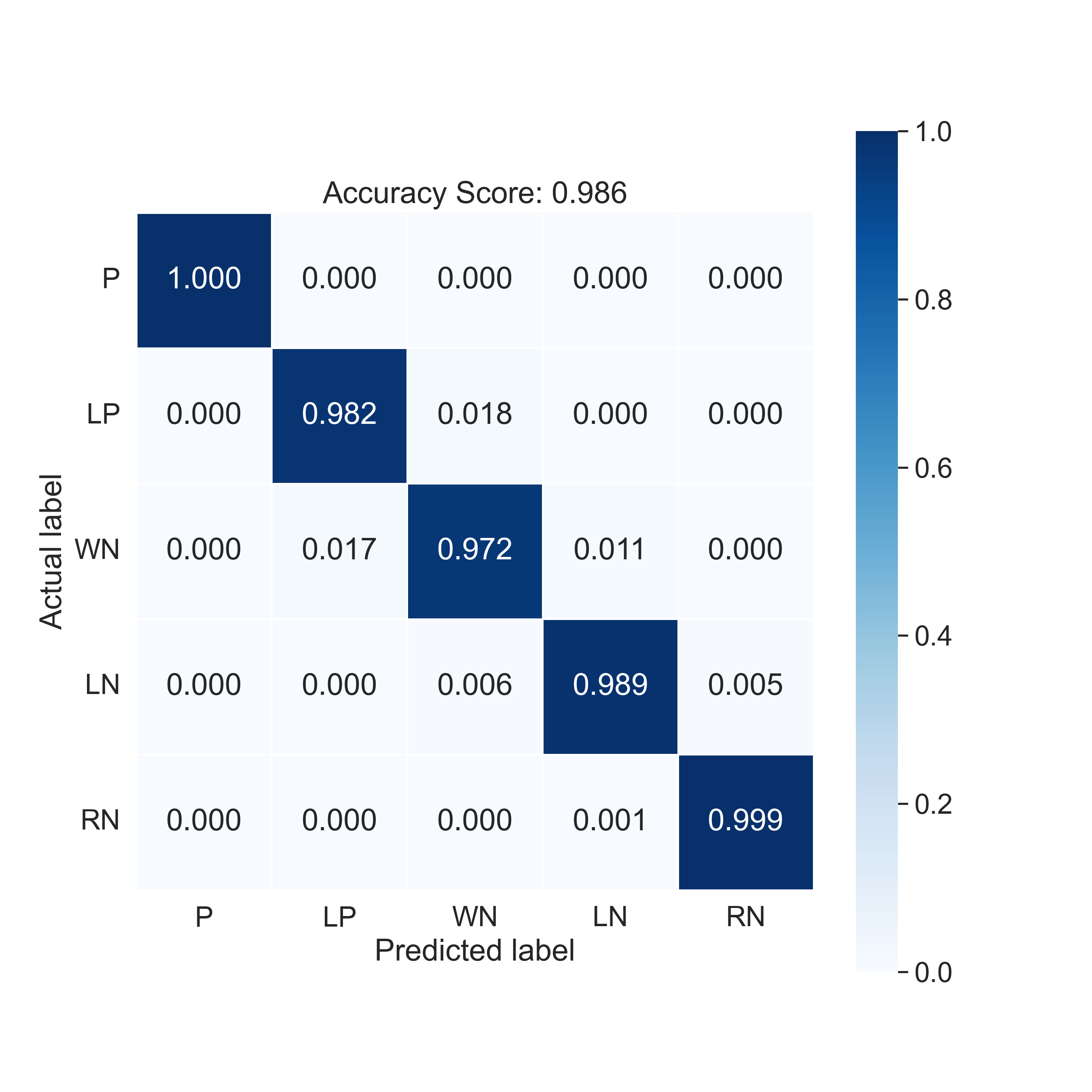}}
    
    \subfigure[SVM + TFO features]{\includegraphics[width=0.32\textwidth]{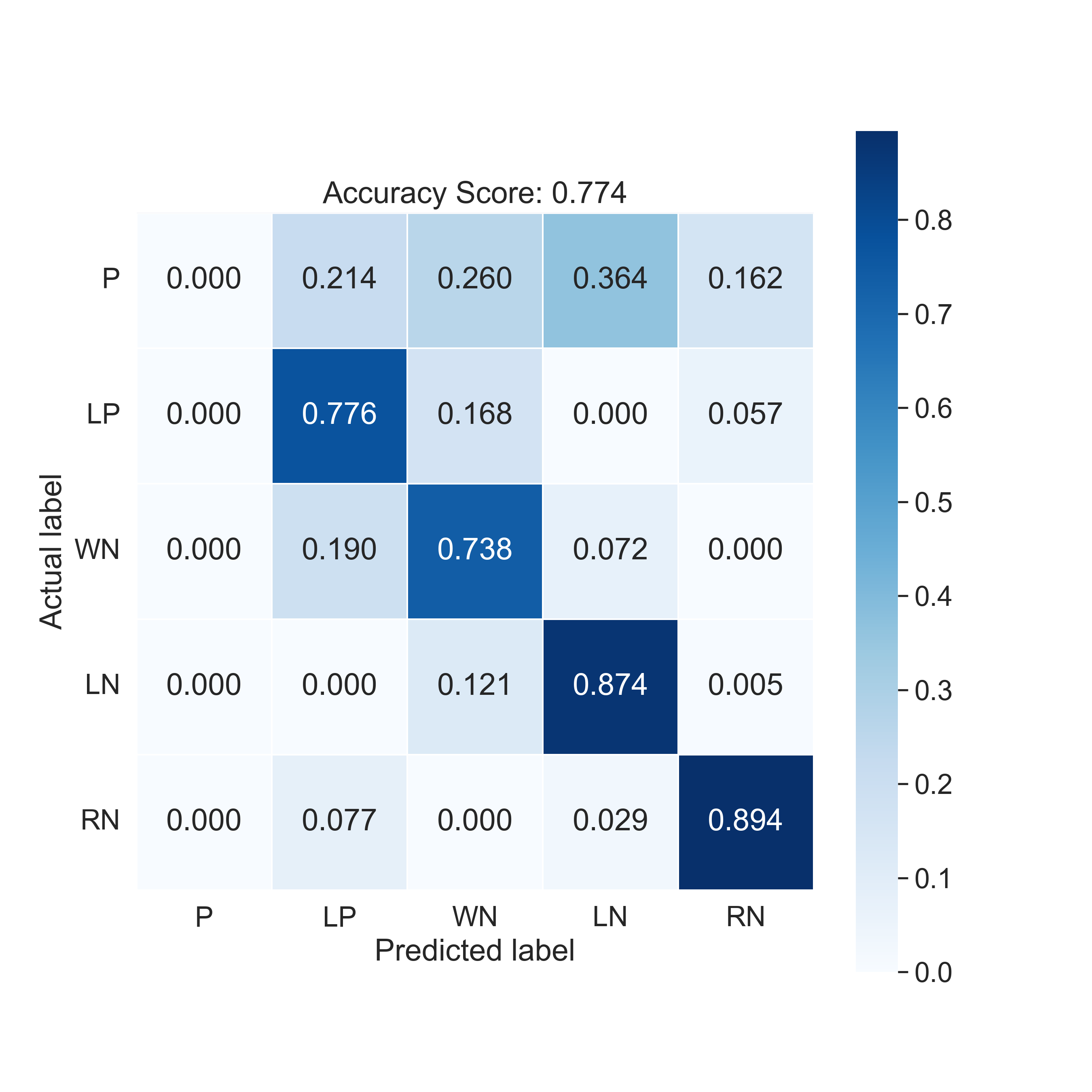}} 
    \subfigure[SVM + PUDI features]{\includegraphics[width=0.32\textwidth]{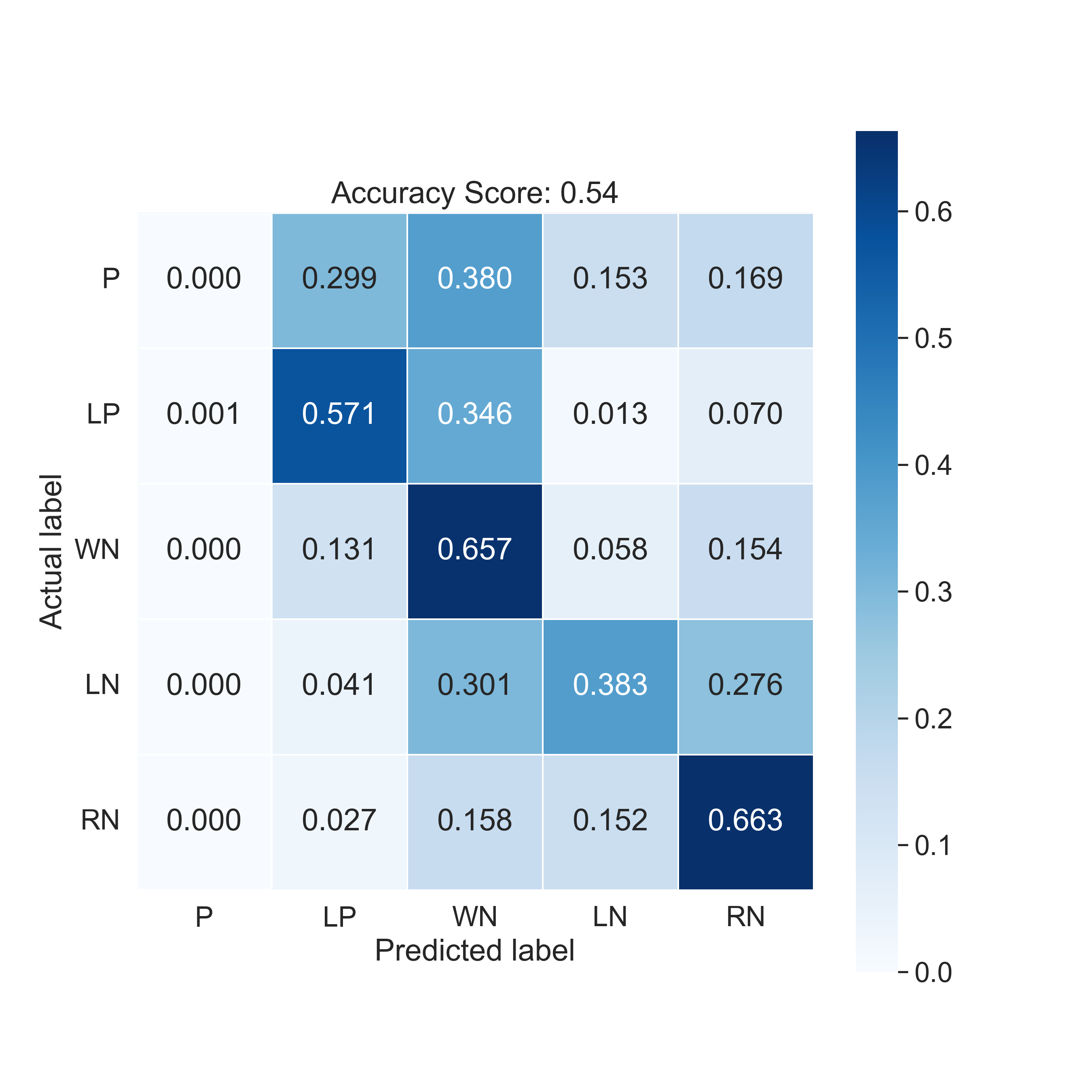}} 
    \subfigure[SVM + NeDBIT features]{\includegraphics[width=0.32\textwidth]{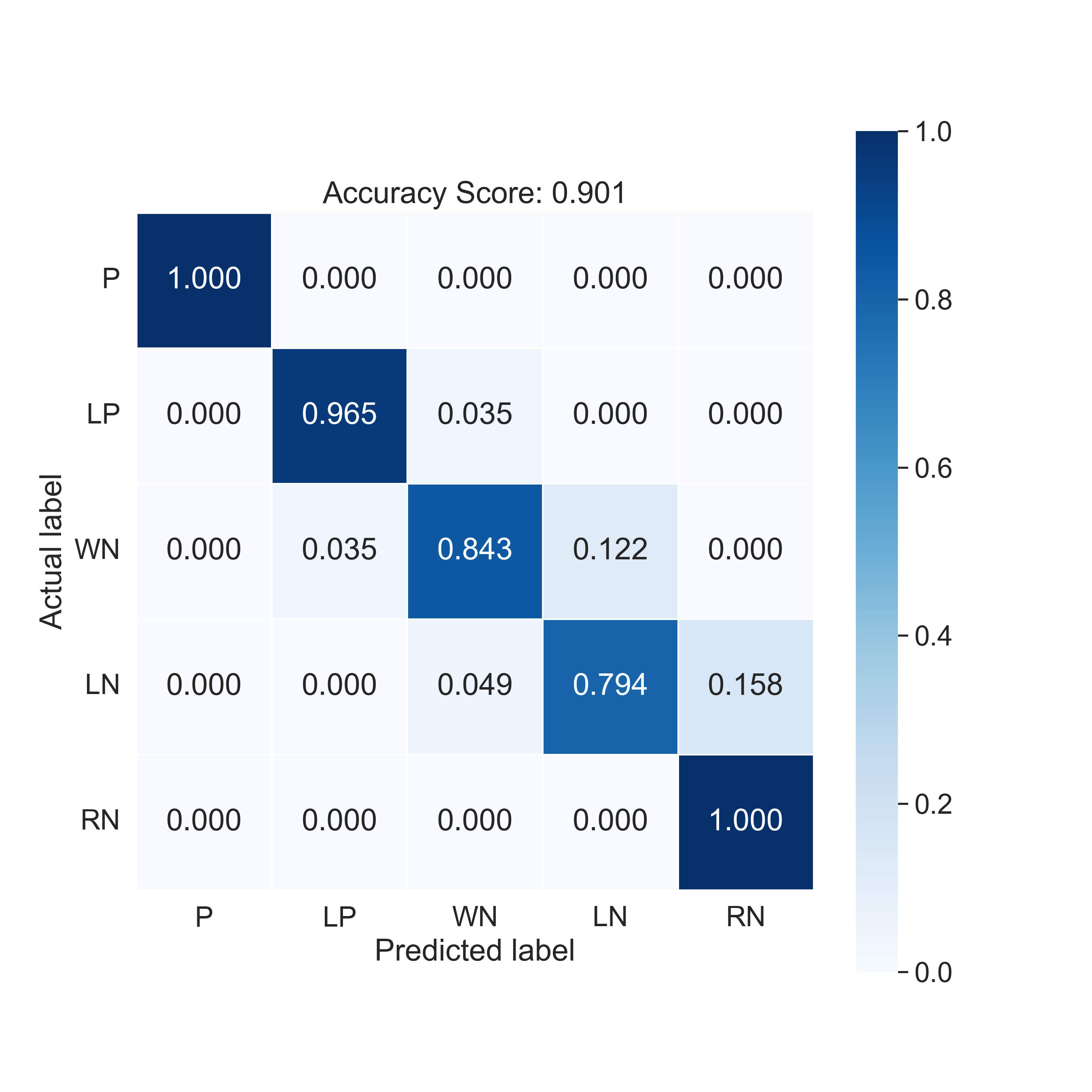}}
    \caption{Confusion matrices for multi-class classification on malignant neoplasm of breast (C0006142). The APU labeling and the newly defined NeDBIT features allow for a better and clear distinction of the P class and the pseudo-classes.}
    \label{fig:multi-class}
\end{figure*}

Results related to malignant neoplasm of breast disease are reported in Figure~\ref{fig:multi-class} in terms of confusion matrices.  
The comparison among TFO, PUDI, and NeDBIT features shows that the latter are far superior to the others. The joint usage of APU and NeDBIT features (NIAPU) succeeded in discriminating the class P from the rest of the genes and better separating the pseudo-classes LP, WN, LN, and RN.

Regarding the pseudo-classes, the identification performances were also satisfying using TFO and PUDI features, even if with a drop in accuracy compared to NeDBIT. This highlights the effectiveness of the APU label assignment. RF and MLP delivered the best performances. Regarding SVM, LN samples were sometimes misclassified as either WN or RN.

Overall, for P and RN classes, the NIAPU classification is almost perfect since NeDBIT features allow those classes to be properly separated from the others since they grasp the topological aspects of the set of seed genes as a whole, assigning lower and lower weights to genes that are progressively “far” from the set of seed genes. For the rest of the classes, the performances are good but some genes are misclassified. This is due to the label assignment via quantiles, which obviously introduces some arbitrary noise at the boundary of such quantiles.

Results related to the other diseases are provided in Supplementary File 2, along with the results of a five-fold cross-validation study carried out for the three sets of features.

\subsection{NIAPU performances in disease gene identification}
\label{subsec:LP}

We tested the ability of NIAPU to identify new candidate genes. We performed a validation by excluding the 20\% of seed genes, setting them as unlabeled both in the computation of the NeDBIT features and in the APU labeling algorithm. We repeated the procedure five times with non-overlapping gene sets. We investigated whether NIAPU was able to properly classify the removed positive genes as LP. For brevity, the results for malignant neoplasm of breast only are reported in Table~\ref{tab:lp-APU} (other diseases in Supplementary File 2). 
On average, around 46\% of unlabeled seed genes fell in the LP class, while the rest fell in a decreasing classification trend toward the RN class.
We also observed a clear correspondence between the labeling and the association score: the higher the score, the more likely the gene is to be found in the LP class. This underlines the influence of scores on the NeDBIT features. Analogous results can be found in Supplementary File 2 for the remaining diseases.

Aggregated results related to ML classification for malignant neoplasm of breast are reported in Table~\ref{tab:ML-class-kfold-multi}. 
All the classes were identified by RF and MLP with high scores, while SVM reported lower metrics, particularly with regard to the LN class. Therefore, NIAPU turned out to be robust also in more challenging settings with reduced seed gene sets.

\begin{table}[!ht]
\centering
\caption{Labeling of the unlabeled seed genes by NIAPU for malignant neoplasm of breast (C0006142). Results are intended as average with standard deviation over the five runs (GDAS: association score $\mathbf{S}$).}
\label{tab:lp-APU}
\begin{small}
\noindent\makebox[\columnwidth]{
\begin{tabular}{@{}llllll@{}}
\toprule
label & \% genes       & \# genes     & GDAS mean      & GDAS median    & GDAS mode     \\ \midrule
LP    & 45.659 ± 1.362 & 93.6 ± 2.793 & 0.383 ± 0.016 & 0.346 ± 0.019 & 0.32 ± 0.045 \\
WN    & 27.415 ± 0.636 & 56.2 ± 1.304 & 0.343 ± 0.013 & 0.318 ± 0.011 & 0.3 ± 0.0    \\
LN    & 17.659 ± 4.436 & 36.2 ± 9.094 & 0.324 ± 0.012 & 0.303 ± 0.004 & 0.3 ± 0.0    \\
RN    & 9.268 ± 3.65   & 19.0 ± 7.483 & 0.322 ± 0.013 & 0.303 ± 0.004 & 0.3 ± 0.0    \\ \bottomrule
\end{tabular}
}
\end{small}
\end{table}

\begin{table}[!ht]
\centering
\caption{Classification scores as pooled mean and standard deviation (over all the diseases). Five runs were performed for each disease, masking out 20\% of seed genes.}
\label{tab:ML-class-kfold-multi}
\begin{tabular}{@{}llll@{}}
\toprule
label        & precision       & recall          & F1 score        \\ \midrule
\textbf{MLP} &                 &                 &                 \\
P            & 0.994 ± 0.011 & 0.998 ± 0.007 & 0.996 ± 0.007 \\
LP           & 0.972 ± 0.013 & 0.972 ± 0.016 & 0.972 ± 0.012 \\
WN           & 0.955 ± 0.02  & 0.915 ± 0.022 & 0.933 ± 0.019 \\
LN           & 0.835 ± 0.021 & 0.744 ± 0.042 & 0.782 ± 0.019 \\
RN           & 0.731 ± 0.037 & 0.86 ± 0.036  & 0.788 ± 0.024 \\
macro avg    & 0.898 ± 0.008 & 0.898 ± 0.007 & 0.894 ± 0.008 \\
weighted avg & 0.884 ± 0.009 & 0.876 ± 0.009 & 0.876 ± 0.009 \\ \midrule
accuracy     & 0.876 ± 0.009 &                 &                 \\ \midrule
\textbf{RF}  &                 &                 &                 \\
P            & 1.0 ± 0.0     & 1.0 ± 0.0     & 1.0 ± 0.0     \\
LP           & 0.984 ± 0.005 & 0.984 ± 0.005 & 0.984 ± 0.005 \\
WN           & 0.977 ± 0.007 & 0.976 ± 0.007 & 0.977 ± 0.006 \\
LN           & 0.982 ± 0.005 & 0.986 ± 0.004 & 0.984 ± 0.004 \\
RN           & 0.991 ± 0.003 & 0.987 ± 0.004 & 0.989 ± 0.003 \\
macro avg    & 0.987 ± 0.003 & 0.987 ± 0.003 & 0.987 ± 0.003 \\
weighted avg & 0.984 ± 0.004 & 0.984 ± 0.004 & 0.984 ± 0.004 \\ \midrule
accuracy     & 0.984 ± 0.004 &                 &                 \\ \midrule
\textbf{SVM} &                 &                 &                 \\
P            & 0.998 ± 0.004 & 1.0 ± 0.0     & 0.999 ± 0.002 \\
LP           & 0.845 ± 0.043 & 0.719 ± 0.071 & 0.767 ± 0.032 \\
WN           & 0.635 ± 0.135 & 0.726 ± 0.108 & 0.625 ± 0.102 \\
LN           & 0.625 ± 0.191 & 0.559 ± 0.026 & 0.419 ± 0.025 \\
RN           & 0.366 ± 0.224 & 0.5 ± 0.004   & 0.38 ± 0.011  \\
macro avg    & 0.694 ± 0.066 & 0.701 ± 0.013 & 0.638 ± 0.022 \\
weighted avg & 0.641 ± 0.077 & 0.642 ± 0.017 & 0.568 ± 0.029 \\ \midrule
accuracy     & 0.642 ± 0.017 &                 &                 \\ \bottomrule
\end{tabular}
\end{table}

\subsection{NIAPU vs. other disease gene identification tools}
\label{sec:gene-discovery}
We compared the predictive performance in the identification of candidate disease genes of NIAPU against known gene discovery algorithms, namely DIAMOnD \citep{pmid25853560}, Markov clustering (MCL) \citep{pmid11917018, sun2011prediction}, random walk with restart (RWR) \citep{kohler2008, valdeolivas2019random}, two variants of GUILD \citep{guney2012exploiting}, one exploiting the NetCombo measure and the other based on Functional Flow (fFlow) \citep{nabieva2005whole}, and ToppGene \citep{chen2009disease} (relying on the implementation provided by the GUILD software). See Supplementary File 2 for a detailed description of these algorithms.
For this analysis, we relied on the extended GDA dataset provided by DisGeNET. We assigned the labels using NIAPU on the curated version of the dataset and then investigated whether the seed genes contained in the extended version (but not in the curated one) fell into the LP class. We considered the ranking retrieved by NIAPU at different quantile thresholds.

\begin{figure*}[!htbp]
    \centering
    \subfigure[Malignant neoplasm of breast (C0006142)]{\includegraphics[width=0.49\textwidth]{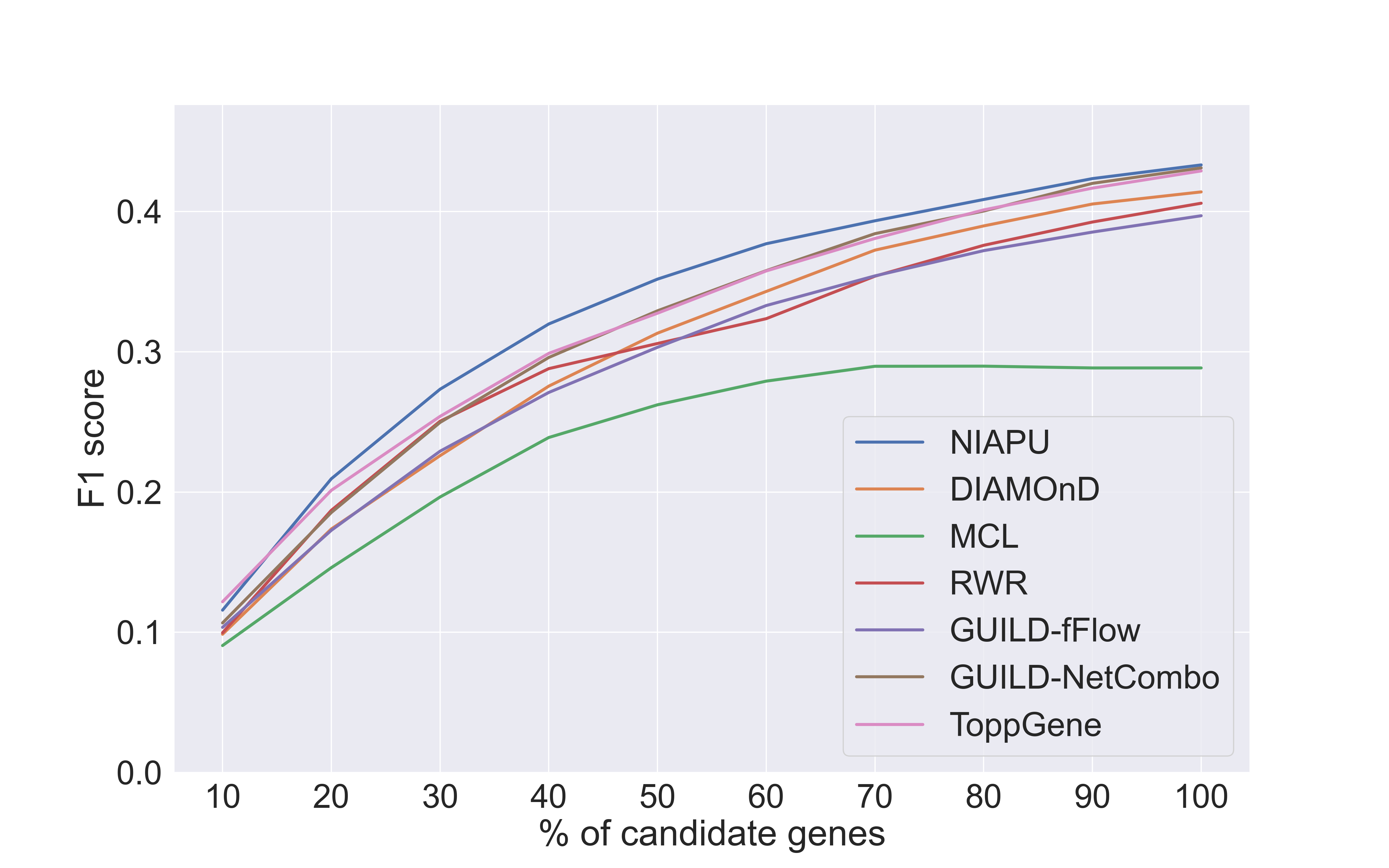}} 
    \subfigure[Schizophrenia (C0036341)]{\includegraphics[width=0.49\textwidth]{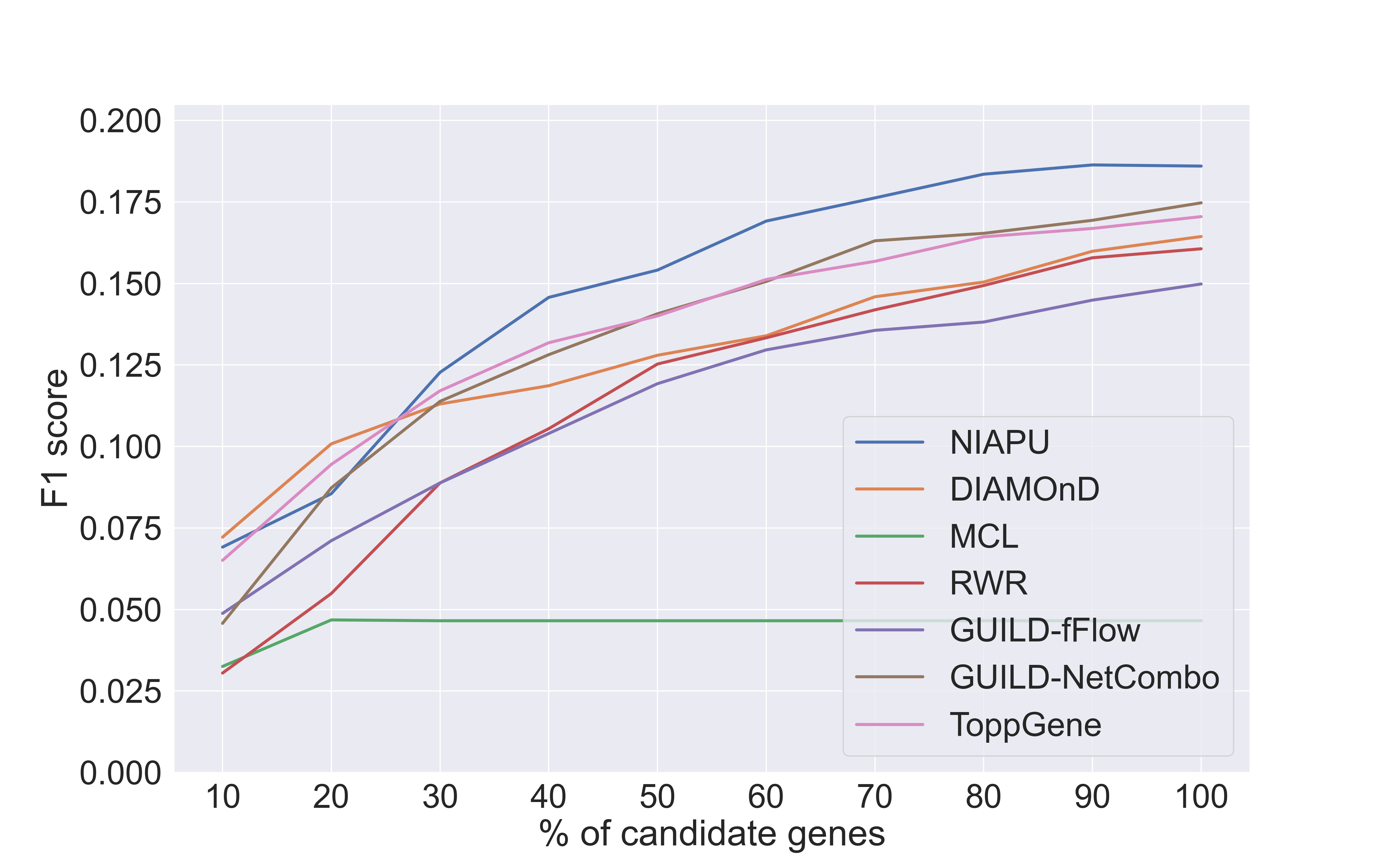}} \\
    
    \subfigure[Colorectal carcinoma (C0009402)]{\includegraphics[width=0.49\textwidth]{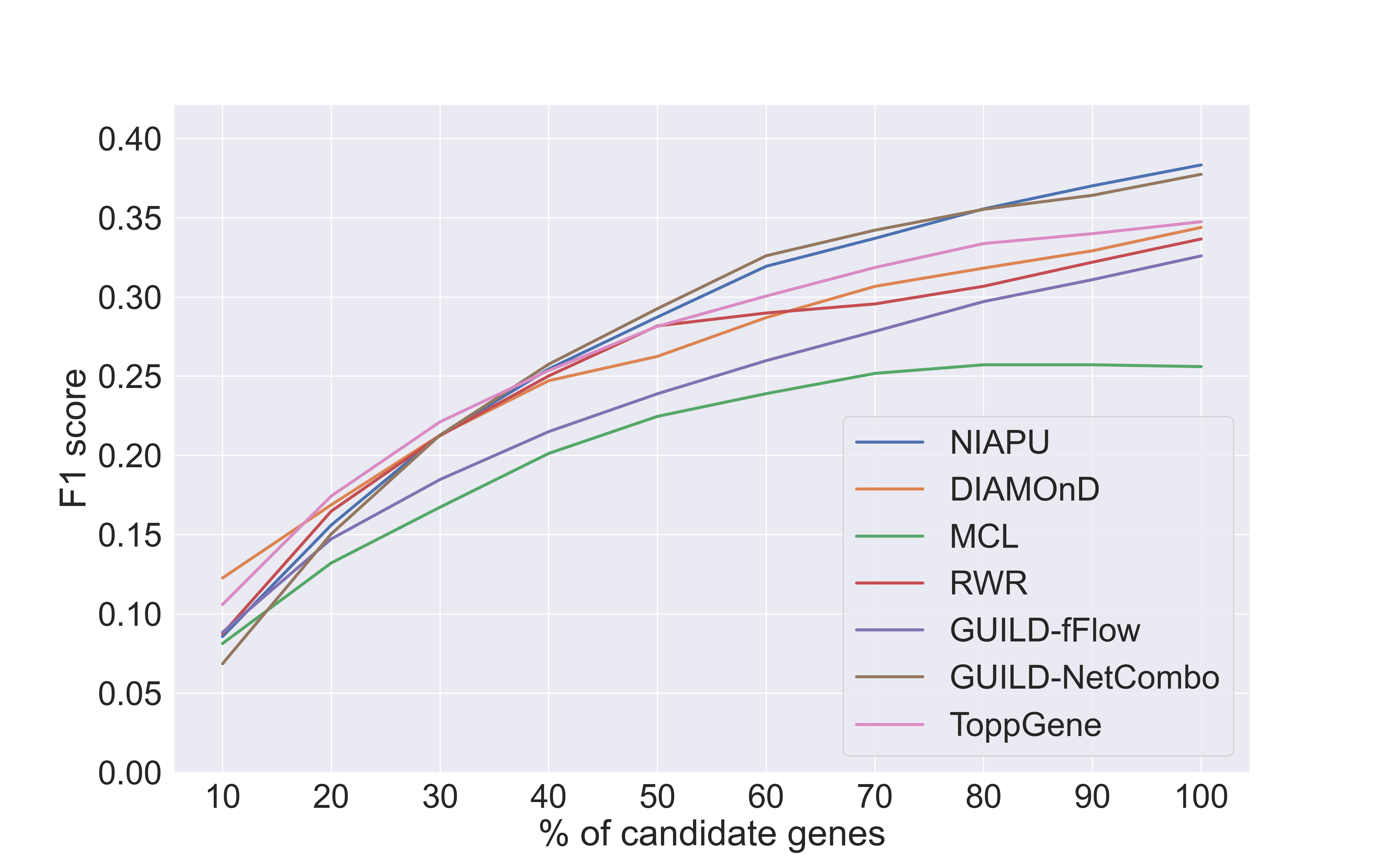}} 
    \subfigure[Malignant neoplasm of prostate (C0376358)]{\includegraphics[width=0.49\textwidth]{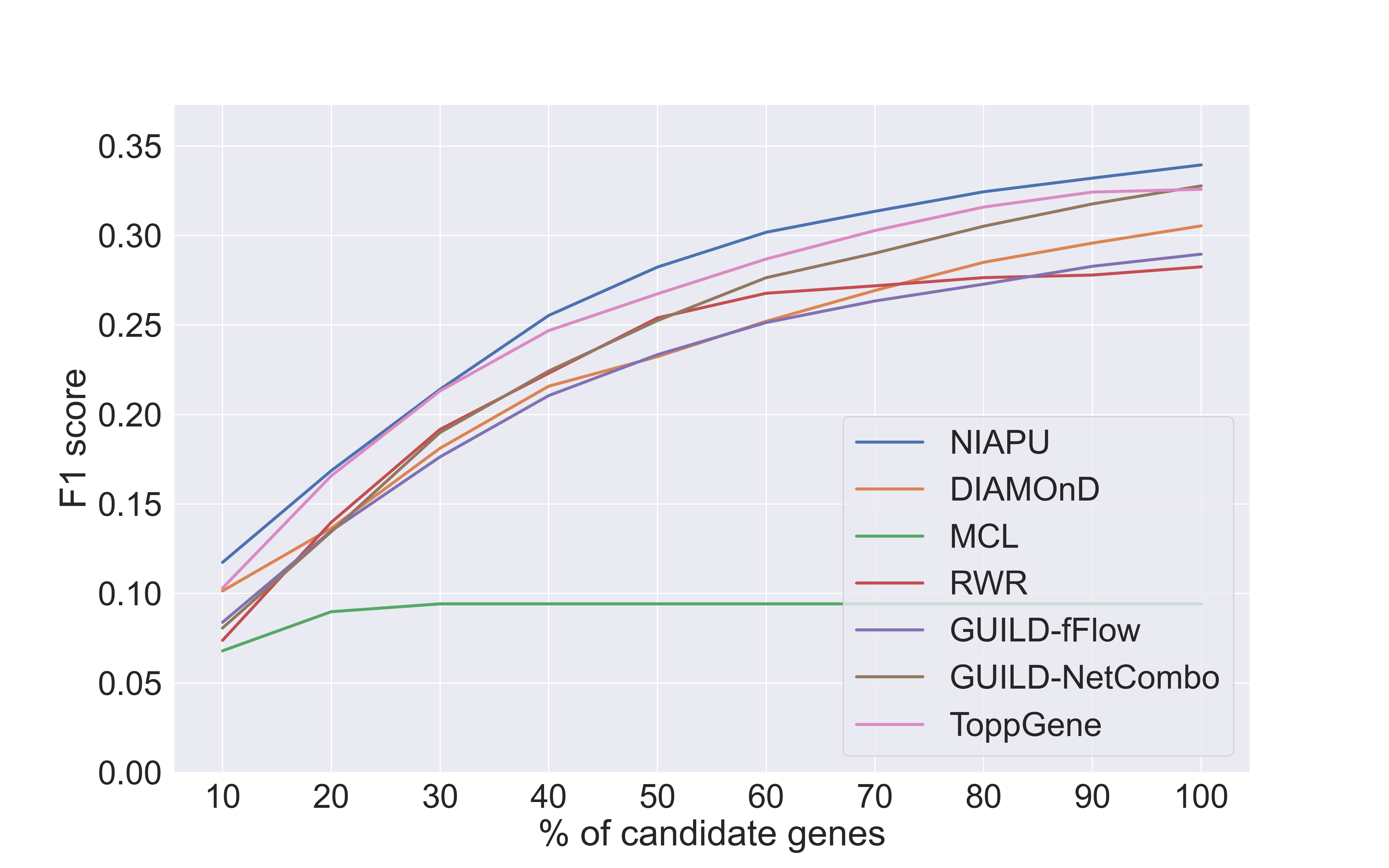}}\\
    
    \subfigure[Bipolar disorder (C0005586)]{\includegraphics[width=0.49\textwidth]{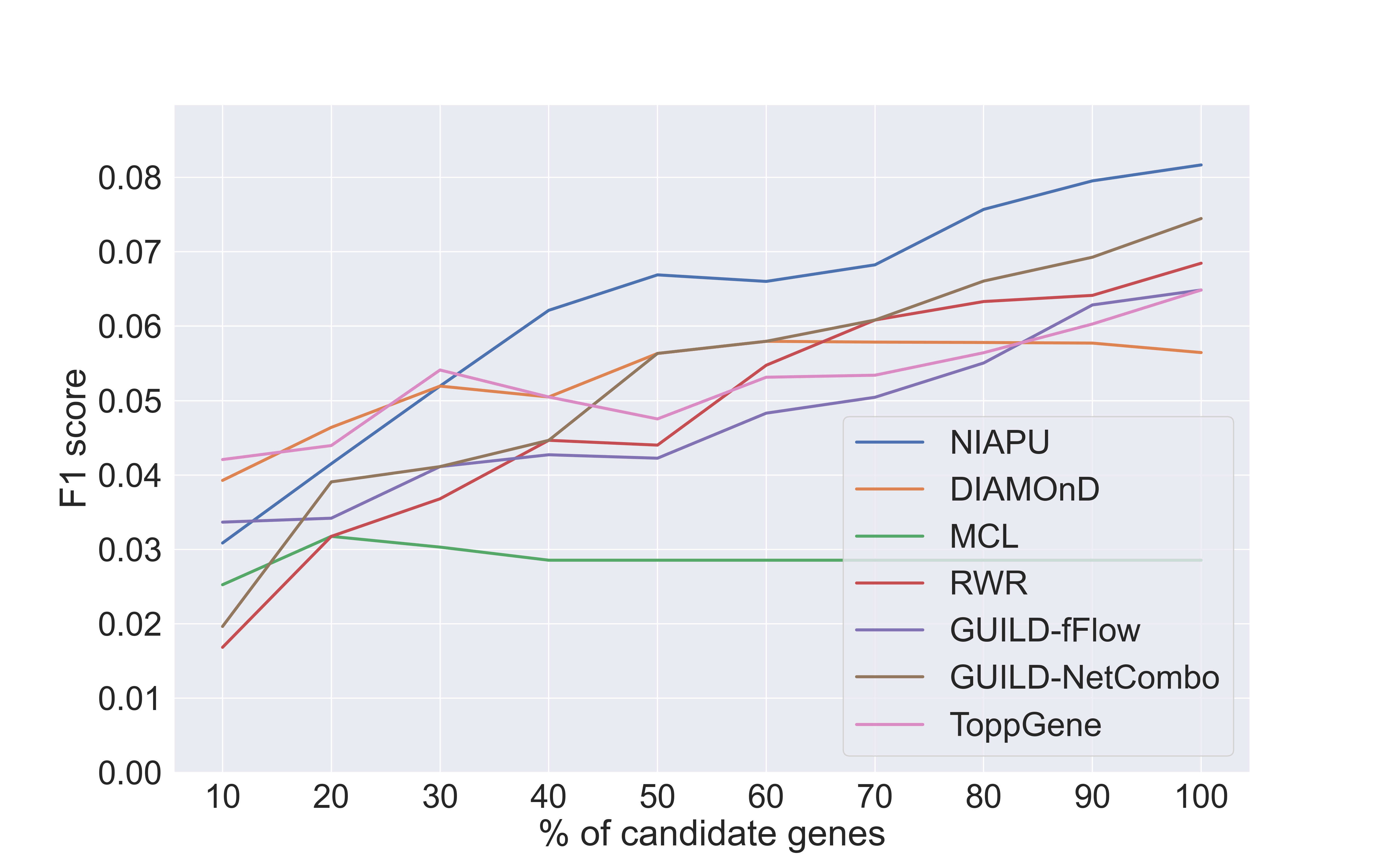}} 
    \subfigure[Depressive disorder (C0011581)]{\includegraphics[width=0.49\textwidth]{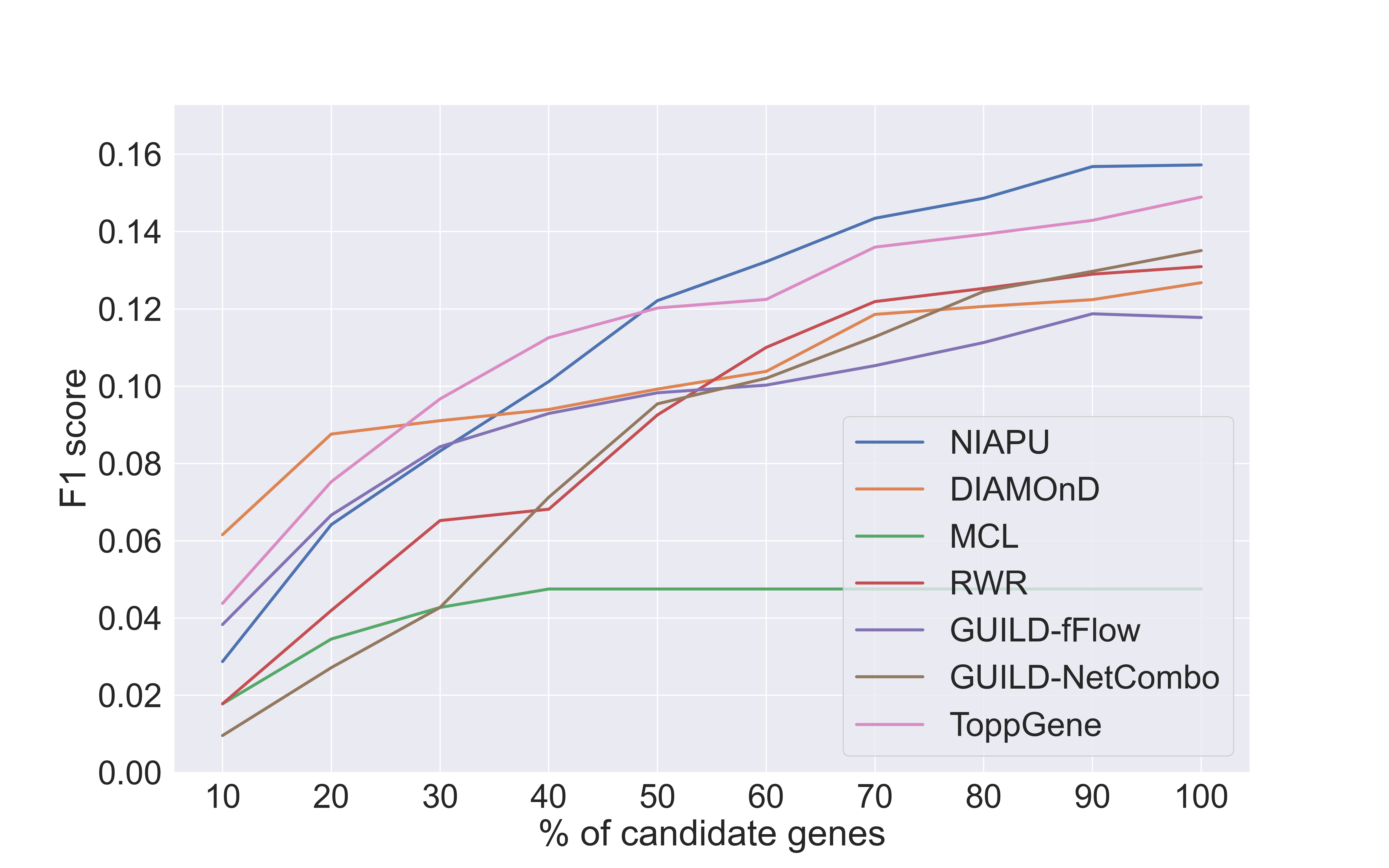}}\\
   
    \caption{Gene discovery performances in terms of F1 score. Results are reported for six diseases for increasing numbers of candidate genes considered as a percentage of the total number of associated genes in the extended dataset, which is different for each disease. The rest of the diseases can be found in Supplementary File 2.}
    \label{fig:extented-validation}
\end{figure*}

In Figure~\ref{fig:extented-validation}, we report the results of this comparison in terms of F1 score. Most of the time, our methodology outperformed or was at par with the state-of-the-art algorithms for disease gene identification, being often the best-performing method when looking for a large number of candidate genes and of comparable performances for lower ones. Indeed, DIAMOnD performs at its best when considering a low ratio (10-20\%) of predicted genes, while NIAPU shows good performances both for low and high percentages of candidate genes, outperforming DIAMOnD in the latter case. In fact, as stated by the authors themselves, DIAMOnD becomes unreliable when exceeding 200 predicted genes \citep{pmid25853560}.

\subsection{Enrichment analysis}\label{sec:enrichment}

For a further evaluation of our results, for each of the ten diseases considered, we performed a gene ontology/pathway/disease enrichment analysis of the first 100 predicted genes in the LP class from the validation on the extended GDA dataset. This analysis was performed using Enrichr \citep{chen2013enrichr, kuleshov2016enrichr, enrichr}.

The selected LP genes do not correspond to any of the curated GDA disease genes; therefore, among the enriched diseases, we cannot expect to find the same disease for which the gene discovery process is carried out. Instead, among the enriched terms (diseases, GO terms, or pathways), we should be able to find diseases and biological processes that are somehow related to the disease under scrutiny.

We report the enrichment analysis results in Table~\ref{tab:enrichment}. In particular, we present the top enriched diseases or biological processes for each analyzed disease, together with references to literature that endorse such relevant links.

Although not conclusive, the fact that there is evidence in literature of links and shared biological mechanisms between the analyzed diseases and enriched diseases is additional proof of the validity and efficacy of the disease gene discovery process.

\begin{table*}[ht]
\centering
\caption{Enrichment analysis of the LP genes predicted for the ten diseases of interest. The top enriched diseases and GO terms are reported, along with notes about disease relationships and main reference articles.}
\label{tab:enrichment}
\noindent\makebox[\columnwidth]{
\begin{tabular}{@{}llll@{}}
\toprule
\textbf{Disease}                                                                      & \textbf{Enriched disease/GO}                                                                           & \textbf{Relationship}                                                                                                                                                                                                                                  &  \textbf{Reference} \\ \midrule
\begin{tabular}[c]{@{}l@{}}C0036341\\ Schizophrenia\end{tabular}                       & \begin{tabular}[c]{@{}l@{}}KEGG \\ GO:0042981\\ Regulation of\\ apoptotic processes\end{tabular}       & \begin{tabular}[c]{@{}l@{}}Apoptotic engulfment pathway \\ involved in schizophrenia \\ (increased risk)\end{tabular}                                                                                                                           &  \citealp{pmid19721717}       \\ \midrule
\begin{tabular}[c]{@{}l@{}}C0005586\\ Bipolar \\ disorder (BD)\end{tabular}                 & \begin{tabular}[c]{@{}l@{}}KEGG \\ GO:0042981   \\ Regulation of \\ apoptotic processes\end{tabular}   & \begin{tabular}[c]{@{}l@{}}Observed relationship between\\ mitochondrial dynamics \\ and dysfunction and the \\ apoptotic pathway activation \\ and the pathophysiology of BD\end{tabular}                                                      &  \citealp{pmid28463235}       \\ \midrule
\begin{tabular}[c]{@{}l@{}}C0006142\\ Malignant \\ neoplasm of\\ breast\end{tabular}   & Leukaemia                                                                                              & \begin{tabular}[c]{@{}l@{}}Therapy-related myeloid neoplasms \\ may be part of a cancer-risk syndrome \\ involving breast cancer\end{tabular}                                                                                                              &  \citealp{pmid22220266}       \\ \midrule
\begin{tabular}[c]{@{}l@{}}C0009402\\ Colorectal \\ carcinoma \\  (CRC)\end{tabular}             & Ovarian cancer (OC)                                                                                             & \begin{tabular}[c]{@{}l@{}}GCNT3 might constitute a prognostic \\ factor also in OC \\ and emerges as an essential \\ glycosylation-related molecule \\ in CRC and OC progression\end{tabular}                                                &  \citealp{pmid29855486}       \\ \midrule
\multicolumn{1}{c}{}                                                                   & Parkinson                                                                                              & \begin{tabular}[c]{@{}l@{}}Neurobiological investigations suggest\\ that depression in Parkinson's \\ disease may be mediated by \\ dysfunction in mesocortical/prefrontal \\ reward, motivational, and \\ stress-response systems\end{tabular} & \citealp{pmid1372794}        \\
\begin{tabular}[c]{@{}l@{}}C0011581\\ Depressive\\ disorder\end{tabular}               &                                                                                                        &                                                                                                                                                                                                                                                 &                                       \\
                                                                                       & \begin{tabular}[c]{@{}l@{}}GO:0043066 \\ Negative regulation \\ of apoptotic \\ processes\end{tabular} & \begin{tabular}[c]{@{}l@{}}Evidence of local inflammatory, \\ apoptotic, and oxidative stress \\ in major depressive disorder\end{tabular}                                                                                                      & \citealp{pmid20479761}      \\ \midrule
\begin{tabular}[c]{@{}l@{}}C0023893\\ Liver\\ cirrhosis\end{tabular}                   & Parkinson                                                                                              & \begin{tabular}[c]{@{}l@{}}Parkinson's disease among the \\ neurological complications in advanced \\ liver cirrhosis mediated by manganese\end{tabular}                                                                                                                                               &  \citealp{pmid33072457}      \\ \midrule
\begin{tabular}[c]{@{}l@{}}C0376358\\ Prostate \\ cancer\end{tabular}                  & Melanoma                                                                                               & \begin{tabular}[c]{@{}l@{}}Diagnoses of cutaneous melanoma \\ may be associated with prostate \\ cancer incidence\end{tabular}                                                                                                                  & \citealp{pmid29740153}       \\ \midrule
\begin{tabular}[c]{@{}l@{}}C3714756\\ Intellectual \\ disability\end{tabular}          & Dementia                                                                                               & \begin{tabular}[c]{@{}l@{}}People with intellectual disability are \\ at higher risk of dementia \\ than the general population\end{tabular}                                                                                                    & \citealp{zigman2007alzheimer}      \\ \midrule
                                                                                       & Ovarian cancer (OC)                                                                                             & \begin{tabular}[c]{@{}l@{}}Alcohol consumption might be associated \\ with the risk of OC in specific   \\ populations or in studies \\ with specific characteristics\end{tabular}                                                  & \citealp{yan2015association}      \\
\begin{tabular}[c]{@{}l@{}}C0860207\\ Chronic \\ alcoholic\\ intoxication\end{tabular} &                                                                                                        &                                                                                                                                                                                                                                                 &                                       \\
                                                                                       & \begin{tabular}[c]{@{}l@{}}KEGG Estrogen \\ signaling pathway\end{tabular}                             & \begin{tabular}[c]{@{}l@{}}Association of increased estrogen level and\\ increased alcohol use in females\end{tabular}                                                                                                                          & \citealp{erol2019sex}       \\ \midrule
\begin{tabular}[c]{@{}l@{}}C0001973\\ Drug-induced \\ liver disease\end{tabular}       & Leigh Syndrome (LS)                                                                                    & \begin{tabular}[c]{@{}l@{}}Valproate, listed as a cause of \\ drug-induced acute liver failure,\\ can cause mitochondrial dysfunction\\ and should be avoided in LS patients\end{tabular}                                                       & \citealp{lee2021clinical}       \\ \bottomrule
\end{tabular}
}
\end{table*}

\section{Discussions and conclusions}
\label{sec:conclu}
In this paper, we presented the NIAPU algorithm, which fits the typical problem of the computational identification of previously unknown disease genes in the context of positive-unlabeled learning. 
The advantage of the proposed method is that it allows accurate characterization of the positive samples (P set) -- via the NeDBIT features -- and refined control of the likely positive samples (LP set) -- via the APU labeling procedure -- which, extracted from the set of unlabeled elements, contains, with the highest probability, elements related to the disease of interest.
Moreover, NIAPU turned out to be an effective labeling procedure, allowing machine learning models to be trained appropriately and deliver highly accurate classification performances.
As for disease gene identification, NIAPU proved to be efficient in two different experiments. In the first one, masking out a subset of seed genes, it turned out that \textasciitilde{}46\% of those fell in the LP class. In the second one, assigning labels using NIAPU on the curated version of the DisGeNET dataset and then searching for the seed genes of the extended version only, the predictive performance of the NIAPU algorithm outperformed or was at par with the state-of-the-art algorithms for disease gene discovery.

It is worth noting that the NeDBIT features are designed to be able to use link-weighted and node-weighted graphs and that, by having increasingly accurate PPIs, we expect increasingly good results from the application of NIAPU. On the other hand, NIAPU methodology is clearly influenced by the reliability of seed genes, the association score assigned to them, and the background network topology (here, the PPI network and its reliability). 

Indeed, GDA datasets may be affected by disease-gene association bias due to the quantity of research on a given disease/trait. In this regard, a recent systematic review \citep{de2021every} demonstrated that 87.7\% of all genes could be associated with cancer. This indicates that given the massive amount of research focused on cancer, which also applies to other types of diseases, the definition “associated with” is to be checked carefully and critically.

The usage of datasets that are as error-free, unbiased, and reliable as possible (e.g., using an interactome validated in the specific pathological context, possibly with weighted PPIs) could potentially improve the classification performance of the method. In this regard, it is worth mentioning that an algorithm with the same theoretical ground of NIAPU has been applied in different contexts (e.g., nephrology, gastroenterology, rare diseases) \citep{shahini2022network, 11573_1631720}, paying particular attention to the selection of seed genes and reference interactomes.

%
%
\clearpage

\section*{Acknowledgements}

We would like to thank Prof. Dr. David B. Blumenthal (FAU, Erlangen-Nürnberg, DE) for insightful comments and suggestions on the first version of this work.

\section*{Funding}

This work has been partially supported by the ERC Advanced Grant 788893 AMDROMA “Algorithmic and Mechanism Design Research in Online Markets”, the EC H2020RIA project “SoBigData++” (871042), the MIUR PRIN project ALGADIMAR “Algorithms, Games, and Digital Markets”, and by the CNR project DIT.AD021.161.001 / Analisi probabilistica di dataset biologici e network dynamics.

\bibliographystyle{natbib}

\bibliography{biblio}

\end{document}